\documentclass[journal,twoside,web]{ieeecolor} \providecommand{\refname}{References}
\usepackage{generic}
\usepackage{cite}
\usepackage{amsmath,amssymb,amsfonts}
\usepackage{algorithmic}
\usepackage{graphicx}
\usepackage{algorithm,algorithmic}
\usepackage{hyperref}
\hypersetup{hidelinks=true}
\usepackage{textcomp}
\usepackage{booktabs}
\usepackage{bm}

\usepackage{comment}
\usepackage[T1]{fontenc}
\usepackage[table]{xcolor}
\usepackage{multirow}
\usepackage{dblfloatfix}

\newcommand{\Secg}{\ensuremath{Z_{\text{ecg}}}}
\newcommand{\Scmr}{\ensuremath{Z_{\text{cmr}}}}
\newcommand{\zecg}{\ensuremath{\mathbf{z}_{\text{ecg}}}}
\newcommand{\zcmr}{\ensuremath{\mathbf{z}_{\text{cmr}}}}
\newcommand{\zpred}{\ensuremath{\mathbf{z}_{\text{pred}}}}

\newcommand{\Uecg}{\ensuremath{U_{\mathrm{ecg}}}}
\newcommand{\Ucmr}{\ensuremath{U_{\mathrm{cmr}}}}
\newcommand{\Sshared}{\ensuremath{S}}
\newcommand{\Stil}{\ensuremath{\tilde{S}}}
\newcommand{\Util}{\ensuremath{\tilde{U}_{\mathrm{ecg}}}}
\newcommand{\R}{\ensuremath{\mathbb{R}}}

\def\BibTeX{{\rm B\kern-.05em{\sc i\kern-.025em b}\kern-.08emT\kern-.1667em\lower.7ex\hbox{E}\kern-.125emX}}
\begin{document}
\bstctlcite{IEEEexample:BSTcontrol}
\title{CardioJEPA: A Framework for Holistic Cardiac Representations Through Predictive Multimodal Alignment with ECG-Preservation

\author{Daniel Uyterlinde, Laura Alvarez-Florez, Samuel Ruipérez-Campillo, \IEEEmembership{Member, IEEE}, Lukas P. A. Arts, Folkert W. Asselbergs, Erik J. Bekkers, Ivana Išgum and Fleur V. Y. Tjong }

\thanks{Manuscript submitted for review on September 18th, 2026. This research has been conducted using the UK Biobank Resource under Application Number 24711. This work was supported by grants awarded to F.V.Y.T. by the Dutch Heart Foundation (Dekker 03-005-2025-0129), the Dutch Research Council (NWO Rubicon 452019308 and NWO AINed XS NGF.1609.243.045), Amsterdam Cardiovascular Sciences, and Health~Holland, as well as by the Dutch Research Council (NWO) through the Vidi project SIGN (VI.Vidi.233.220), awarded to E.J.B.}

\thanks{D. Uyterlinde and E. J. Bekkers are with the Informatics Institute, University of Amsterdam, The Netherlands. (e-mail: d.j.uijterlinde@amsterdamumc.nl; e.j.bekkers@uva.nl).}

\thanks{L. Alvarez-Florez is with the Department of Biomedical Engineering and Physics, Amsterdam University Medical Center, The Netherlands, and with the Quantitative Healthcare Analysis Group, Informatics Institute, University of Amsterdam, The Netherlands. (e-mail: l.alvarezflorez@amsterdamumc.nl)}

\thanks{S. Ruipérez-Campillo is with the Department of Computer Science, ETH Zurich, Zurich, Switzerland. (e-mail: samuel.ruiperezcampillo@inf.ethz.ch)}

\thanks{L.P.A. Arts, F.W. Asselbergs, and F.V.Y. Tjong are with the Heart Center, Department of Clinical and Experimental Cardiology, Amsterdam University Medical Centers, Amsterdam Cardiovascular Sciences, University of Amsterdam, Amsterdam, The Netherlands. (e-mail: l.p.a.arts@amsterdamumc.nl, f.w.asselbergs@amsterdamumc.nl, f.v.tjong@amsterdamumc.nl)} 

\thanks{I. I\v{s}gum is with the Department of Radiology, Mayo Clinic, Rochester, MN, USA, the Department of Biomedical Engineering and Physics, Amsterdam University Medical Center, The Netherlands, and with the Department of Radiology and Nuclear Medicine, Amsterdam University Medical Center, The Netherlands. (e-mail: Isgum.Ivana@mayo.edu)}

\thanks{D. Uyterlinde and L. Alvarez-Florez contributed equally to this work as co-first authors. I. I\v{s}gum and F. V. Y. Tjong supervised this work jointly.}}

\maketitle

\begin{abstract}
The electrocardiogram (ECG) is the most widely used cardiac diagnostic tool, but it reflects cardiac structure and function only indirectly. Aligning ECG and cardiac magnetic resonance (CMR) representations offers a way to transfer imaging-derived structural knowledge into the ECG encoder without compromising the electrophysiological information it already carries. Contrastive learning has become the predominant approach for ECG--CMR alignment, yet the quality of the learned representation is jointly determined by the CMR target representation, the alignment objective, and the preservation of electrophysiological information unique to the ECG. We propose CardioJEPA, a predictive alignment framework built on frozen cardiac foundation models that addresses all three: a target that separates static anatomy from cardiac motion, a predictive objective that maps ECG representations onto a fixed CMR target without instance discrimination, and a preservation term that retains nearly all pre-alignment electrophysiological information. On 63,193 paired ECG--CMR recordings from the UK Biobank, CardioJEPA outperforms previous ECG--CMR alignment methods across structural, electrical, and diagnostic tasks, retains 98.6\% of the ECG-unique information present before alignment, substantially reduces the modality gap relative to matched contrastive alignment, and transfers to external cohorts for structural heart disease and Chagas cardiomyopathy without task-specific training, achieving state-of-the-art performance.
\end{abstract}


\section{Introduction}
\label{sec:introduction}
\IEEEPARstart{N}{o} single cardiac modality fully describes the heart. The electrocardiogram (ECG) captures electrical activity over time, transthoracic echocardiography (TTE) visualizes cardiac motion and function, and cardiac magnetic resonance (CMR) provides detailed structural, functional, and tissue-level information. Together, these modalities offer a holistic understanding of cardiac health. Yet, access to all modalities within the same clinical study is rare. Additionally, while ECG is inexpensive and widely available, modalities such as CMR require costly equipment, specialized infrastructure, and trained personnel~\cite{pennell2010cardiovascular, kligfield2007recommendations}. Recent advances in deep learning could offer a solution through multimodal representation learning, which leverages relationships between complementary modalities~\cite{chen2020simclr, ding2024cross}. 
In particular, these modalities can be integrated through \textit{cross-modal alignment}, a process that updates unimodal encoders to produce similar representations in a shared latent space~\cite{alvarezflorez2026dualphasecrossmodalcontrastivelearning, turgut2025ecgcmr}. Crucially, because only the ECG is typically available at clinical deployment, our primary goal is to maximize the diagnostic utility of the ECG representation. We aim to embed CMR-derived structural knowledge into the ECG encoder without compromising the electrophysiological information it already contains, reducing reliance on costly multimodal examinations.

Contrastive learning has become the predominant approach for aligning ECG and CMR representations, following its success in other domains~\cite{chen2020simclr, radford2021clip}. Its effectiveness, however, is not guaranteed; the quality of the resulting ECG representation depends on three key factors: (i) the information encoded in the CMR target, (ii) the objective used to transfer this information to the ECG encoder, and (iii) the extent to which the ECG encoder retains its own modality-specific information during alignment. First, the CMR target bounds what can be transferred at all: alignment cannot recover anatomical or functional characteristics that are poorly represented or absent from it. Second, the objective shapes how that information is transferred. Most previous methods have relied on InfoNCE~\cite{oord2019representationlearningcontrastivepredictive}, which frames alignment as an instance-discrimination task by pulling matched ECG--CMR pairs together while pushing apart all other patients in the batch. Because every unmatched patient contributes to the denominator, phenotypically similar patients are treated as negatives despite sharing clinically meaningful characteristics, distorting the geometry of the learned representation. Third, alignment acts on an ECG encoder that already carries clinically valuable information. In optimizing for cross-modal agreement, the encoder may discard the electrophysiological characteristics that give the ECG independent diagnostic value, and no contrastive formulation includes a mechanism to prevent this.
Consequently, an effective multimodal framework must explicitly retain this modality-unique information during the alignment process.


As a solution, we propose CardioJEPA, a predictive framework for ECG--CMR representation alignment built on pretrained cardiac foundation models. CardioJEPA makes two methodological contributions. First, because raw CMR embeddings are dominated by anatomical variance, we apply PCA independently to the static and motion components of the CMR latent space, giving the functional signal its own subspace and enriching the information available for transfer. Second, CardioJEPA replaces contrastive instance discrimination with latent prediction following the JEPA principle~\cite{lecun2022path,assran2023jepa}. Rather than relying on negative samples, the ECG encoder directly predicts the corresponding CMR representation, which remains frozen throughout alignment. Because only the ECG encoder is updated, it is free to retain information beyond what is shared. A dedicated preservation objective further reinforces this by retaining the electrophysiological information encoded by the original ECG model.

Next, we review related work in Section~\ref{sec:related-work}, then formalize the problem and introduce CardioJEPA in Section~\ref{sec:method}. In Section~\ref{sec:experiments}, we show that the PCA-based CMR target and preservation objective are key determinants of downstream performance, that the unimodal initialization already surpasses prior alignment methods trained from scratch, and that CardioJEPA learns a more comprehensive cardiac representation than competing methods. In the same section, we examine the geometry of the aligned latent space and validate transfer to external cohorts for structural heart disease and Chagas disease detection. We end with a discussion and conclusions in Sections~\ref{sec:discussion} and~\ref{sec:conclusion}, respectively.

\section{Related Work}
\label{sec:related-work}
\subsection{Representation Learning for ECG and CMR}    
Deep learning based analysis of ECG signals has progressed from CNNs that extract local features~\cite{ribeiro_automatic_2020} and CNN--LSTM hybrids that incorporate rhythm-level temporal modeling~\cite{oh2018automated}, toward self-supervised transformer architectures that remove the dependence on large annotated corpora~\cite{gedon2021towardssslecg, mehari2021sslecg}. Scaling self-supervision to millions of signal recordings yields ECG foundation models, whose defining property is label efficiency. Vaid et al.\ \cite{vaid2023foundational} showed that a masked-modeling ViT pretrained on 8.5M ECGs needs an order of magnitude less fine-tuning data than supervised CNNs, a result replicated across objectives~\cite{coppola2024hubert, li2025ecgfounder}.

Following a similar trend, CMR representation learning has shifted from supervised segmentation~\cite{ronneberger2015unet} to self-supervision, via contrastive~\cite{chaitanya2020contrastive} and masked autoencoder~\cite{He_2022_MAE} pretext tasks. Because adjacent regions in cine sequences are highly correlated, masked autoencoders can exploit this spatial redundancy to infer missing content from visible context, forcing the network to learn genuine cardiac anatomy without manual annotations.

\begin{figure*}
    \centering
    \includegraphics[width=0.85\linewidth]{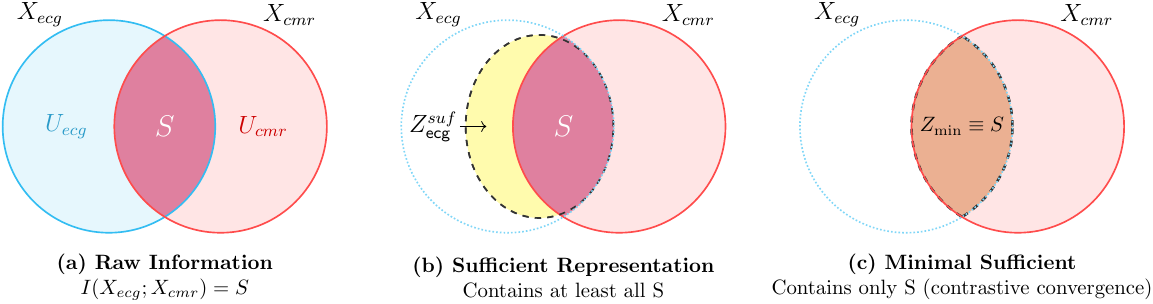}
    \vspace{-0.5em}
    \caption{Information-theoretic diagrams of (minimal) sufficient representations. Contrastive learning (c) encourages minimal sufficiency, discarding ECG-unique information ($\Uecg$). Predictive alignment (b) targets an unconstrained sufficient representation, preserving modality-unique information.}
    \label{fig:info_diagrams}
\end{figure*}

\subsection{Contrastive ECG-CMR Alignment}
The availability of paired ECG and CMR acquisitions for the same individual makes contrastive alignment directly applicable to cardiac multimodal learning. Paired ECG and CMR recordings from the same patient naturally define positive pairs, with all other patients in the batch serving as negatives. 

Early joint ECG-CMR representation learning explored cross-modal autoencoder frameworks, which learn a shared representation by reconstructing both modalities from a common latent space \cite{radhakrishnan2023cross}. Recent methods have moved increasingly toward contrastive alignment, both ECCL \cite{ding2024cross} and MMCL \cite{turgut2025ecgcmr} apply a CLIP-style symmetric InfoNCE objective to patient-matched ECG-CMR pairs, typically following separate unimodal pre-training of each encoder through self-supervised objectives such as masked auto-encoding \cite{He_2022_MAE}. Building on this paradigm, PTACL \cite{selivanov2026PTACL} augments global patient-level alignment with a local temporal contrastive loss that aligns ECG segments with corresponding CMR frames, while \cite{alvarezflorez2026dualphasecrossmodalcontrastivelearning} align ECGs with 3D CMR volumes at end-diastole and end-systole through a dual-phase contrastive loss using frozen CMR encoders, incorporating phase-specific structural information into ECG representations.

Despite its empirical success, the InfoNCE-based contrastive objective common in previous works carries three structural limitations in the clinical multimodal setting. First, instance discrimination treats every patient pair as negative, yet in a predominantly healthy cohort a large fraction are physiologically close neighbors, so pushing them apart scatters clinically similar patients across the manifold. Techniques such as debiased sampling~\cite{chuang2020debiased} and CMR-similarity soft-weighting~\cite{ding2026generating, selivanov2026PTACL, khosla2020supcon} only partially address what is fundamentally a binary approximation to a continuous physiological spectrum. Second, maximizing the mutual-information bound drives the ECG embedding toward a minimal sufficient statistic of the CMR embedding~\cite{oord2019representationlearningcontrastivepredictive, wang2022rethinking}, leaving little capacity for the conduction and rhythm information that gives the ECG independent diagnostic value; we formalize this in Section~\ref{sec:contrastive-limitations}. Third, InfoNCE jointly enforces alignment of paired representations and uniformity over the hypersphere~\cite{wang2020hypersphere}. Uniformity improves instance discrimination, but where the CMR target space already carries a clinically meaningful geometry it overrides that structure, driving the two encoders into narrow, disjoint cones. Liang et al. term this the modality gap~\cite{liang2022mindgapunderstandingmodality}, and the effect is amplified in ECG--CMR alignment by the asymmetry between temporal ECG signals and volumetric (3D+time) CMR representations.

\subsection{Preserving Modality-Unique Information}
\label{sec:decomp_predictive}
The assumptions underlying contrastive learning are well suited to settings where positive pairs correspond to complementary views of the same underlying signal~\cite{tian2020infomin, lee2021predicting, tosh2021contrastive}. ECG and CMR, however, are not two views of the same signal, but complementary measurements of the heart that capture its electrophysiological and structural state. Consequently, electrophysiological characteristics such as P-wave duration have no direct counterpart represented in CMR imaging, meaning that an objective designed to suppress modality-specific information inevitably discards clinically relevant ECG features that are essential for cardiac diagnosis~\cite{liang2023factorized}. This observation motivates the explicit preservation of ECG-unique information as a core component of our alignment framework.

Predictive latent modeling offers an alternative to contrastive alignment. Rather than discriminating positive and negative pairs, a predictor maps a context representation to a target representation. The Joint Embedding Predictive Architecture (JEPA)~\cite{lecun2022path, assran2023jepa} formalizes this in the self-supervised unimodal setting where a predictor maps visible context to the representations of masked regions produced by an exponential moving average (EMA) encoder. Predicting in latent space rather than pixel space lets the model target semantic structure and avoids penalizing it for failing to reconstruct low-level noise. The same principle naturally extends to cross-modal learning. CrossJEPA~\cite{perera2025crossjepa} replaces the EMA teacher with a fixed encoder of a second modality, demonstrating that a predictor can transfer knowledge across modalities against a frozen target, leaving the source encoder free to retain information beyond what is shared. Because the objective is a per-sample regression loss rather than a global repulsion signal, it imposes no uniformity pressure and exerts no direct force to suppress modality-unique variance. Wang et al. \cite{wang2022rethinking} further prove that explicitly maximizing the input--representation mutual information $I(X_{\text{ecg}}; Z_{\text{ecg}})$ as a preservation regularizer reduces both Bayes classification and minimum regression error relative to the minimal sufficient baseline. These observations motivate our approach of combining a predictive cross-modal alignment objective against a frozen CMR target with an explicit term designed to preserve ECG-unique information.
\section{Methodology}\label{sec:method}
This section presents \textbf{CardioJEPA}, our framework for predictive ECG--CMR representation alignment. We first formalize the information an ECG-only representation must carry at deployment and show why contrastive alignment places a ceiling on it. We then describe the four components of the framework: pretrained unimodal encoders, a CMR target that isolates function from anatomy, a predictive alignment objective, and a distillation term that preserves ECG-unique information.  
\textit{Code will be made publicly available.}


\subsection{Problem Formulation}\label{sec:problem}
\subsubsection{Information Decomposition of the ECG–CMR Pair} 
At deployment, only the ECG is available, so the quantity to maximize is the task-relevant information carried by the ECG representation alone. We decompose this quantity to identify which of its components alignment can transfer and which it must preserve.

Let $\Secg$ and $\Scmr$ denote the ECG and CMR encoder outputs and $Y$ a set of downstream clinical phenotypes. Following~\cite{liang2023factorized}, the task-relevant information decomposes as:
\begin{equation}
\label{eq:info_decomp}
\begin{split}
    I(\Secg,\Scmr;Y) &=
    \underbrace{I(\Secg;\Scmr;Y)}_{S:\;\text{Shared}}
     + \underbrace{I(\Scmr;Y\mid\Secg)}_{\Ucmr:\;\text{CMR-unique}}\\
    &\,+ \underbrace{I(\Secg;Y\mid\Scmr)}_{\Uecg:\;\text{ECG-unique}} .
\end{split}
\end{equation}
Here, $I(A;B)$ denotes mutual information, $I(A,B;C)$ the information the joint variable $(A,B)$ carries about $C$, $I(A;B\mid C)$ conditional mutual information, and $I(A;B;C)$ the interaction information, i.e., the information that $A$ and $B$ share about $C$. Following Liang \textit{et al.}~\cite{liang2023factorized}, we identify this quantity as the shared component~$\Sshared$. The three terms are nonnegative and partition the total task-relevant information by the chain rule (Fig.~\ref{fig:info_diagrams}a).

By construction, $\Ucmr$ is inaccessible at deployment. Therefore, the objective reduces to maximizing $I(\Secg;Y)=\Sshared+\Uecg$. The shared component $\Sshared$ comprises attributes observable in both modalities, such as ventricular volumes, ejection fraction, and myocardial mass; $\Uecg$ captures electrophysiological characteristics absent from CMR, including conduction delays and rhythm abnormalities. An effective alignment framework must therefore maximize the shared information transferred through the CMR target, which upper-bounds what can be transferred at all, while preserving $\Uecg$ during alignment (Fig.~\ref{fig:info_diagrams}a). \\

\subsubsection{Information and Geometric Costs of Alignment}\label{sec:contrastive-limitations}
\begin{figure}[t]
    \centering
    \includegraphics[width=0.9\linewidth]{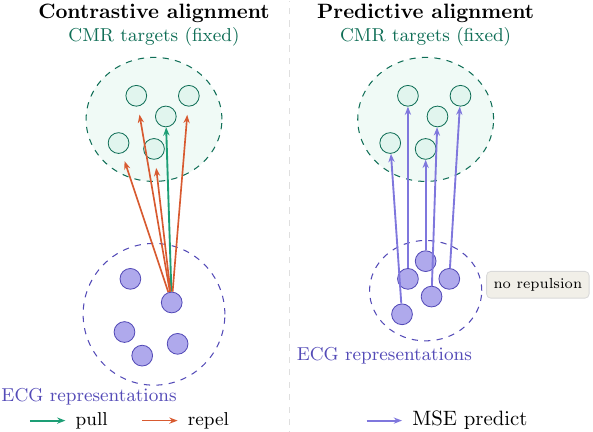}
    \caption{\textbf{Contrastive vs. predictive alignment against a frozen target.} Contrastive alignment (left) introduces batch-wide repulsion that scatters ECG representations. Predictive alignment (right) regresses each ECG representation toward its paired CMR target without repulsion, letting the ECG cluster mirror the geometry of the frozen CMR space.}
    \label{fig:clpred}
\end{figure}

Deployment requires carrying $\Uecg$ through alignment, but InfoNCE instead maximizes a lower bound on the mutual information between the two representations, $I(\Secg;\Scmr)$~\cite{oord2019representationlearningcontrastivepredictive}. This places a ceiling on the ECG-unique information that can be retained:
\begin{align}
    \Uecg = I(\Secg;Y\mid\Scmr) &\leq h(\Secg\mid\Scmr) \nonumber \\
    &= h(\Secg)-I(\Secg;\Scmr),
\label{eq:uecg_ceiling}
\end{align}
where $h(\cdot)$ denotes differential entropy, so $h(\Secg\mid\Scmr)$ quantifies the variability in the ECG representation that the CMR representation does not account for. Asymptotically, InfoNCE drives the encoder toward the minimal sufficient point at which $h(\Secg)$ collapses to $I(\Secg;\Scmr)$ (Fig.~\ref{fig:info_diagrams}c), so $h(\Secg \mid \Scmr) \to 0$ and $\Uecg$ vanishes with it, leaving $\Sshared$ as the only task-relevant information the representation retains~(Fig.~\ref{fig:info_diagrams}c)~\cite{wang2022rethinking,tian2020infomin}. A second limitation is geometric: the InfoNCE gradient weights each negative by its softmax similarity, so repulsion is strongest between the most phenotypically similar patients, precisely where similarity is clinically meaningful (Fig.~\ref{fig:clpred}).\\

\subsubsection{Decoupling Shared and Modality-Unique Information}
With finite encoder capacity, increasing $\Sshared$ appears to come at the expense of $\Uecg$, but only if the conditional entropy $h(\Secg\mid\Scmr)$ is left unconstrained. Contrastive alignment drives $h(\Secg \mid \Scmr)$ toward zero; predictive alignment does not penalize variance orthogonal to the target, but offers no mechanism to retain it either~(Fig.~\ref{fig:info_diagrams}b). Because the ECG foundation model captures substantial electrophysiological information before alignment, $h(\Secg\mid\Scmr)$ starts from a non-trivial baseline: the challenge is not to learn more ECG-unique information, but to prevent its suppression as the encoder specializes toward $\Scmr$. This motivates maximizing
\begin{equation}
\mathcal{J}(\theta)=I(\Secg;\Scmr)+\alpha\,h(\Secg\mid\Scmr),
\label{eq:objective}
\end{equation}
which decouples shared from unique information: $\alpha$ preserves pretrained electrophysiological variance rather than requiring new capacity. Section~\ref{sec:preservation} lower-bounds this term by anchoring to the pretrained representation, instantiating it as a \textit{distillation loss}.

 \begin{figure*}[t]
    \centering
    \includegraphics[width=0.95\linewidth]{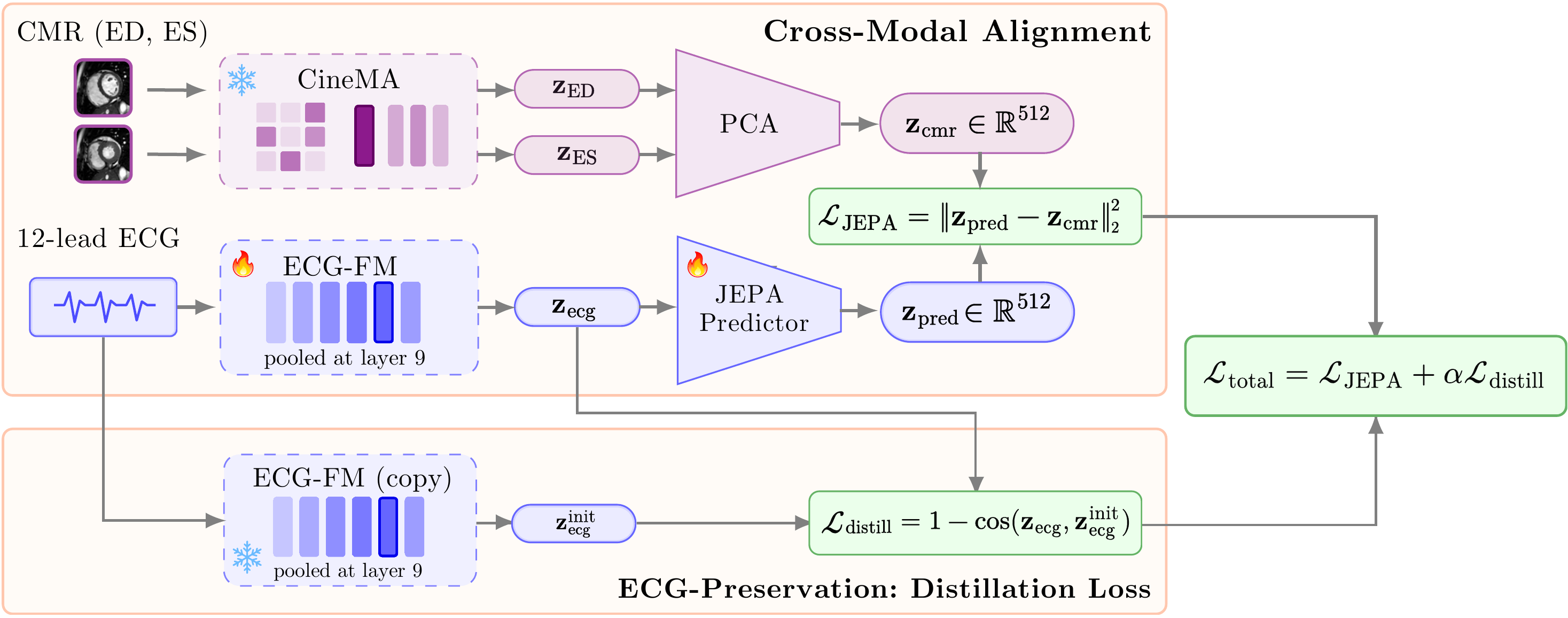}
    \caption{\textbf{CardioJEPA architecture.} The framework aligns an ECG encoder to a frozen, PCA-reduced CMR target ($\mathbf{z}_\text{cmr}$) using a predictive mean-squared error loss ($\mathcal{L}_{\text{JEPA}}$). A distillation loss ($\mathcal{L}_{\text{distill}}$) anchors the trainable ECG representation ($\mathbf{z}_\text{ecg}$) to its frozen initialization ($\mathbf{z}_\text{ecg}^{\text{init}}$), preserving ECG-unique information.}
    \label{fig:overview}
    \vspace{-0.5em}
\end{figure*}

\subsection{Pretrained Encoders and Data Preprocessing}
We initialize the ECG and CMR encoders from the frozen unimodal foundational models ECG-FM~\cite{mckeen2025ecgfm} and CineMA~\cite{fu2025cinema}, respectively. Crucially, both encoders natively encode substantial information regarding cardiac phenotypes prior to any cross-modal alignment.

ECG-FM is a CNN-Transformer hybrid pretrained on 1.5 million 12-lead ECGs using a combination of masked reconstruction and contrastive learning. Its robust baseline performance across downstream tasks makes it an optimal starting point, allowing our framework to focus entirely on structural knowledge transfer and electrophysiological preservation. During preprocessing, ECG signals are split into 5-second windows, with one window drawn uniformly at random per forward pass to serve as a temporal augmentation, and independently z-scored per lead. For the ECG encoder, all representations are extracted at layer 9 (either before or after the predictive CMR alignment). Through layer-wise linear probing, we established this as the optimal extraction point for maximizing both structural and electrical task performance.

CineMA is a multi-view masked autoencoder that jointly reconstructs masked frames across short- and long-axis views, yielding embeddings that capture 3D cardiac geometry without segmentation supervision. The publicly available CineMA weights were trained on the entire UK Biobank CMR cohort~\cite{ukbb}. To avoid data leakage during downstream evaluation, we retrained the CineMA architecture from scratch on a strictly isolated 70/10/20 subject split (100 epochs, with a total of 5.32M images).  We extract the end-diastolic (ED) and end-systolic (ES) frames using the pretrained CineMA segmentation model. 
We keep this encoder frozen throughout alignment. Predicting against a fixed target rules out the degenerate solution in which predictor and target encoder jointly collapse $Z_\text{cmr}$ to a constant, a failure mode of joint-embedding objectives that do not rely on negative pairs~\cite{assran2023jepa}. Beyond this, large-scale self-supervised pre-training has already given the encoder a strong representation of cardiac anatomy and motion, and fine-tuning it on the alignment objective alone risks degrading representations that are already well suited to the task. Prior ECG--CMR alignment work adopts the same asymmetry~\cite{ding2024cross, alvarezflorez2026dualphasecrossmodalcontrastivelearning}.
 
\subsection{Creating a CMR Target Representation}\label{sec:targets}
The CMR target representations must capture both static anatomy and dynamic function. We find that CineMA ED and ES embeddings exhibit a mean cosine similarity exceeding 0.996 on the training split. Because the raw ED and ES embeddings are nearly identical, the alignment loss is largely explained by their shared anatomical component, leaving only a weak training signal for cardiac function. To address this imbalance, we construct an unsupervised, PCA-reduced target, which we call the \emph{isolated-PCA} target, that explicitly separates function from anatomy to prevent static anatomical variance from overshadowing the subtle functional signal. We apply PCA independently to the ED embedding (static anatomy) and to the motion vector $\bm{\Delta}=\mathbf{z}_{\mathrm{cmr}}^{\mathrm{ES}}-\mathbf{z}_{\mathrm{cmr}}^{\mathrm{ED}}$ (function), $z$-score normalize each stream, and concatenate:
\begin{equation}
\zcmr=\big[\,\mathrm{PCA}_{256}(\mathbf{z}_{\mathrm{cmr}}^{\mathrm{ED}})\;\big\|\;\mathrm{PCA}_{256}(\bm{\Delta})\,\big]\in\R^{512},
\end{equation}
where $\big\|$ denotes concatenation and $\text{PCA}_{256}(\cdot)$ the projection onto the top 256 principal components, fitted on the training split. Projecting the two streams separately eliminates the anatomy--motion cross-covariance blocks, preventing anatomical variance from masking the functional signal. Under a Gaussian assumption, the retained variance increases monotonically with $I(Z^{\mathrm{raw}}_{\mathrm{cmr}};Z_{\mathrm{target}})$, so among all rank-$k$ linear projections, PCA is the information-maximizing choice, and the resulting target is the most information-rich linear compression of the CMR state that gives anatomy and function equal weight. This departs from the standard JEPA target, produced by an EMA teacher~\cite{lecun2022path} or, in the cross-modal case, a frozen encoder of the second modality~\cite{perera2025crossjepa}. We retain the frozen-encoder asymmetry but add a fixed linear projection on top of it, a deliberate choice given the near-degeneracy of the raw ED and ES embeddings.

\subsection{The Alignment Objective}\label{sec:objective}
CardioJEPA aligns the ECG encoder to the frozen CMR target (Section~\ref{sec:targets}) through a predictive objective. Our formulation is asymmetric, with gradients flowing only through the ECG branch such that the CMR manifold stays fixed.
A linear predictor $g_\phi$ on top of the ECG encoder predicts the frozen CMR target, $\zpred = g_\phi(\zecg)$:
\begin{equation}
\mathcal{L}_{\mathrm{JEPA}}=\frac{1}{B}\sum_{i=1}^{B}\big\|\zpred^{(i)}-\zcmr^{(i)}\big\|_2^2 .
\end{equation}
The predictor gradient $2g_\phi^{\!\top}(g_\phi\mathbf{z}_\text{ecg}^{(i)}-\zcmr^{(i)})$ has no sum over patients, no softmax weights, and no repulsion: phenotypically similar patients exert zero repulsion, so the ECG cluster can mirror the geometry of the frozen target (Fig.~\ref{fig:clpred}). Because ECG-FM produces $T{=}156$ tokens, we aggregate by token-wise projection with a learned softmax weighting $w^{(t)}$, refining the definition to $\zpred=\sum_t w^{(t)} g_\phi(\zecg^{(t)})$, which emphasizes the most CMR-informative segments. Fig.~\ref{fig:overview} gives an overview.



\subsection{Preserving ECG-Unique Information}\label{sec:preservation}
CardioJEPA pairs the predictive objective with a \textit{distillation loss} that preserves ECG-unique information, giving $\mathcal{L}_\text{total} = \mathcal{L}_\text{JEPA} + \alpha\,\mathcal{L}_\text{distill}$. It acts as a cosine penalty anchoring the active encoder to its pretrained embedding:
\begin{equation}
    \mathcal{L}_{\mathrm{distill}}=\frac{1}{B}\sum_{i=1}^{B}\bigl(1-\cos(\mathbf{z}_{\mathrm{ecg}}^{(i)},\mathbf{z}_{\mathrm{ecg}}^{\mathrm{init},(i)})\bigr),
\end{equation}
where $\mathbf{z}_{\mathrm{ecg}}^{\mathrm{init}}$ is the embedding produced by the frozen copy of the pretrained ECG encoder. The loss penalizes angular deviation only, so the encoder may rescale freely while its angular orientation is held close to the pretrained embedding.

\section{Experiments \& Results}\label{sec:experiments}
We evaluate CardioJEPA by progressively validating the design principles introduced in Section~\ref{sec:method}. We first establish the dataset and evaluation metrics (Sections~\ref{sec:protocol} and~\ref{sec:evaluation}), then assess each design component independently: the CMR target representation (Section~\ref{sec:target}), the alignment and preservation objectives (Section~\ref{sec:paradigm}), and the resulting latent space geometry (Section~\ref{sec:latent}). Finally, we compare CardioJEPA with prior ECG--CMR alignment methods (Section~\ref{sec:positioning}) and evaluate its out-of-distribution generalization (Section~\ref{sec:ood}).

\subsection{Dataset}
\label{sec:protocol}
Our experimental cohort comprises 63{,}193 matched ECG--CMR pairs from the UK Biobank, of which 34{,}044 have available clinical phenotype annotations (Table~\ref{tab:splits}). All participants provided written informed consent as described in the original cohort publications~\cite{ukbb}. From the available UK Biobank annotations, we selected five CMR-derived structural phenotypes: left ventricular (LV) ejection fraction (LVEF), left ventricular end-diastolic volume (LVEDV), left ventricular mass (LVM), right ventricular end-diastolic volume (RVEDV), and left ventricular global longitudinal strain (LVGLS). We additionally selected four ECG-derived phenotypes: PR interval, QRS duration, corrected QT interval (QTc), and P-wave duration, together with four clinical diagnoses: atrial fibrillation (AF), bundle branch block (BBB), cardiomyopathy (CM), and myocardial infarction (MI).
\begin{table}[bhtp]
\centering
\vspace{-0.5em}
\caption{UK Biobank dataset splits.}
\label{tab:splits}
\begin{tabular}{lrl}
\toprule
\textbf{Split} & $N$ & \textbf{Purpose} \\
\midrule
Training  & 47{,}683 & Alignment; probe fit on labeled subset \\
Validation    &  5{,}299 & Alignment loss / early stopping \\
LP-validation &  3{,}404 & In-training probe evaluation \\
LP-test&  6{,}807 & Held-out test \\
\midrule
Total  & 63{,}193 & \\
\bottomrule
\end{tabular}
\end{table}
\subsection{Evaluation Metrics}\label{sec:evaluation}
To quantify both structural transfer and preservation of ECG-unique information, we probe frozen representations using Ridge regression for continuous phenotypes ($R^2$), and logistic regression for binary outcomes (AUROC), reported on the held-out test split, LP-test (Table~\ref{tab:splits}). 

To summarize the trade-off between structural transfer and electrophysiological preservation, we define two composite metrics. $\Stil$ represents the shared information and is computed as the mean $R^2$ across three phenotypes accessible to both modalities: LVEDV, RVEDV, and LV mass. $\Util$ measures ECG-unique information using four electrophysiological phenotypes: PR interval, QRS duration, QTc interval, and P-wave duration. To isolate electrophysiological information beyond cardiac structure, we first remove the component that can be linearly explained by five structural covariates (LVEF, LVEDV, LV mass, RVEDV, and LVGLS) using ordinary least squares (OLS) residualization. The regression coefficients are estimated on the training split and applied to the test split. OLS residuals are orthogonal to these covariates by construction~\cite{frisch1933regression}, so any remaining decodable variance is inaccessible from the five structural summaries used. A positive residual $R^2$ therefore provides direct evidence of linearly accessible electrophysiological information beyond cardiac structure.

\begin{table}[b]
\centering
\vspace{-1em}
\caption{Linear probing $R^2$ of CMR target representations.}
\label{tab:target}
\setlength{\tabcolsep}{4pt}
\begin{tabular}{lcccccc}
\toprule
\textbf{Representation} & \textbf{Dim.} & \textbf{LVEDV} & \textbf{RVEDV} & \textbf{LVM} & \textbf{LVEF} & \textbf{LVGLS} \\
\midrule
\multicolumn{7}{c}{\textit{Raw CineMA embeddings}}\\
ED                           & 768  & \underline{0.954} & 0.939 & \underline{0.965} & 0.247 & 0.322 \\
ES                           & 768  & 0.900 & 0.895 & 0.958 & 0.507 & 0.392 \\
$\bm{\Delta}$ (ES$-$ED)      & 768  & 0.932 & 0.924 & 0.925 & 0.651 & 0.551 \\
Concat (ED, ES)              & 1536 & \textbf{0.965} & \textbf{0.953} & \textbf{0.973} & \textbf{0.691} & \textbf{0.606} \\
\midrule
\multicolumn{7}{c}{\textit{Constructed targets (ours)}}\\
Supervised                   & 256  & \underline{0.954} & \underline{0.942} & 0.964 & 0.628 & 0.565 \\
Isolated PCA  & 512  & \underline{0.954} & 0.939 & 0.960 & \underline{0.663} & \underline{0.570} \\
\bottomrule
\end{tabular}
\end{table}

\begin{figure*}[t]
    \centering
    \vspace{-0.5em}
    \includegraphics[width=0.69\linewidth]{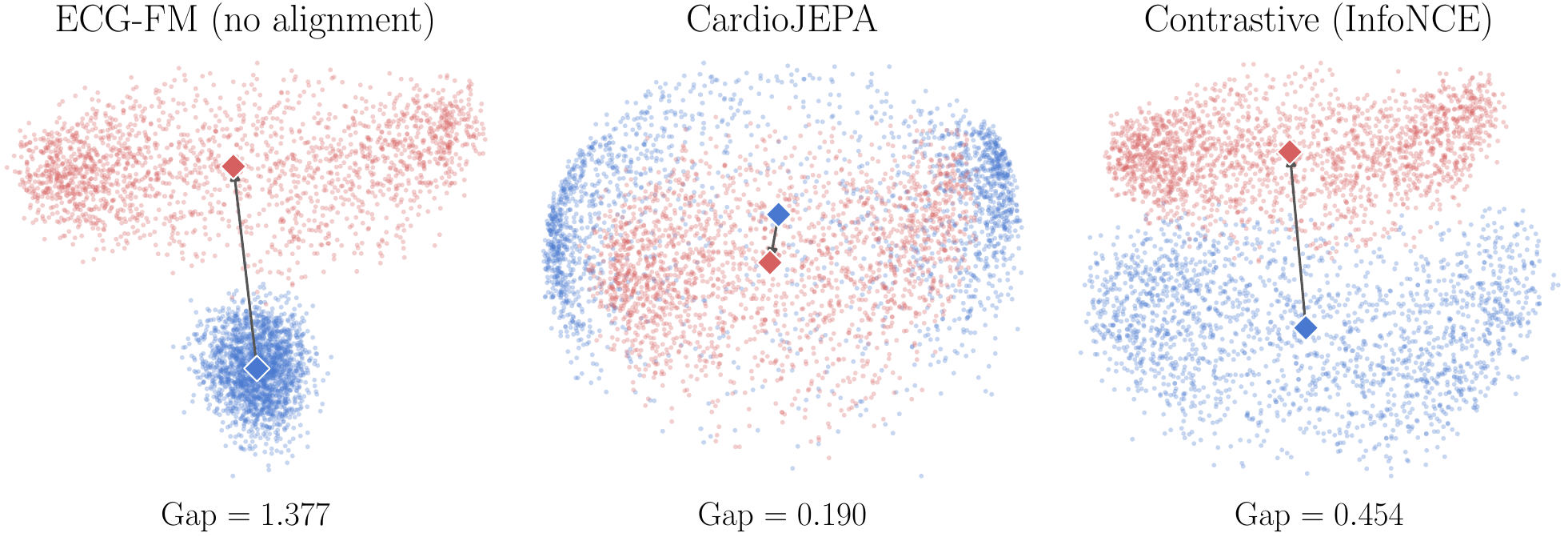} \\
    \includegraphics[width=0.24\linewidth]{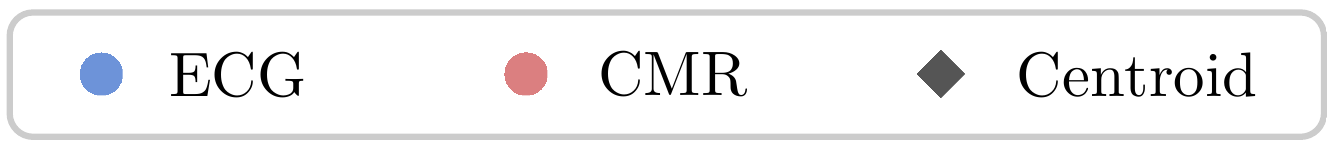}
    \caption{\textbf{Modality gap in the aligned latent space (supervised target).} First two principal components of $\ell_2$-normalized ECG and CMR embeddings on the held-out test set; the (modality) gap is the $\ell_2$ distance between centroids. CardioJEPA achieves clear overlap.}
    \label{fig:gap}
    \vspace{-0.5em}
\end{figure*}
\subsection{The Importance of the CMR Alignment Target}\label{sec:target}
To assess the structural and functional information encoded by different CMR target representations, we evaluate how well each target predicts clinically relevant cardiac phenotypes via linear probing. We compare six target representations: a static end-diastolic (ED) frame, a static end-systolic (ES) frame, their difference ($\bm{\Delta}$) as a motion representation, their concatenation, a supervised representation learned to predict clinical variables, and the proposed isolated-PCA target. The supervised target is constructed by taking the intermediate representations of an MLP, trained to regress/classify 7 cardiac variables on top of the CMR encoder: LVEF, LVM, LVEDV, RVEDV, LVGLS, AF and MI. 
\begin{table}
\centering
\caption{Paradigm comparison at $\alpha=0$ (no preservation term). $\Stil$: structural composite $R^2$. $\Util$: ECG-unique composite $R^2$. Mean $\pm$ std. over three seeds.}
\label{tab:pure}
\begin{tabular}{lcc}
\toprule
\textbf{Method} & $\bm{\tilde{S}}$ & $\bm{\Util}$ \\
\midrule
Predictive (CardioJEPA)           & $0.636\pm0.002$ & $\mathbf{0.660\pm0.008}$ \\
Contrastive & $0.636\pm0.002$ & $0.642\pm0.003$ \\
\bottomrule
\end{tabular}
\end{table}
\begin{table}
\centering
\caption{Alignment cost: $\bm{\Util}$ deviation from the unaligned ECG-FM baseline, by method and preservation setting.}
\label{tab:cost}
\begin{tabular}{lcc}
\toprule
\textbf{Method} & $\bm{\Util}$ ($R^2$) & $\bm{\Delta}$ \textbf{from init} \\
\midrule
ECG-FM~\cite{mckeen2025ecgfm} & 0.709 & --- \\
CardioJEPA + $\mathcal{L}_\text{distill}$ & 0.699 & $-0.010$ \\
Contrastive Learning + $\mathcal{L}_\text{distill}$ & 0.689 & $-0.020$ \\
CardioJEPA ($\alpha=0$) & 0.660 & $-0.049$ \\
Contrastive Learning ($\alpha=0$) & 0.642 & $-0.067$ \\
\bottomrule
\end{tabular}
\end{table} 

\begin{table*}[b!]
\centering
\vspace{-0.5em}
\caption{Linear-probing performance (R\textsuperscript{2} for regression, AUROC for classification) on structural, electrical, and diagnostic cardiac phenotypes, positioned against prior ECG--CMR alignment methods.}
\label{tab:positioning}
\setlength{\tabcolsep}{4.5pt}
\begin{tabular}{llcccc|cccc|cccc|cc}
\toprule
& & \multicolumn{4}{c|}{\textbf{Structural ($R^2$)}} & \multicolumn{4}{c|}{\textbf{Electrical ($R^2$)}} & \multicolumn{4}{c|}{\textbf{Diagnosis (AUROC)}} & \multicolumn{2}{c}{\textbf{Composites ($R^2$)}} \\
\midrule
\textbf{Input} & \textbf{Method} & \textbf{LVEF} & \textbf{LVM} & \textbf{LVEDV} & \textbf{RVEDV} & \textbf{PR} & \textbf{QRS} & \textbf{QTc} & \textbf{P-wave} & \textbf{AFib} & \textbf{BBB} & \textbf{CM} & \textbf{MI} & $\bm{\tilde S}$ & $\bm{\tilde U}_{\text{ecg}}$\\
\midrule

\textbf{ECG} &
\textit{ECG-FM~\cite{mckeen2025ecgfm}}&
\textit{0.286} & \textit{0.649} & \textit{0.554} & \textit{0.559} &
\textit{0.912} & \textit{0.796} & \textit{0.703} & \textit{0.506} &
\textit{0.724} & \textit{0.826} & \textit{0.795} & \textit{0.712} &
\textit{0.587} & \textit{0.709}\\

\midrule

\multirow{4}{*}{\textbf{ECG}} & 
ECCL$^\dagger$~\cite{ding2024cross} &
0.176 & 0.449 & 0.372 & 0.397 &
-- & -- & -- & -- &
-- & -- & -- & -- &
0.406 & --\\

&
MMCL$^\ddagger$~\cite{turgut2025ecgcmr} &
0.139 & 0.401 & 0.342 & 0.363 &
0.494 & 0.640 & 0.488 & 0.252 &
0.683 & 0.767 & 0.629 & 0.689 &
0.369 & 0.474\\

&
PTACL~\cite{selivanov2026PTACL} &
0.257 & 0.613 & 0.505 & 0.531 &
0.872 & 0.770 & 0.657 & 0.406 &
0.719 & \textbf{0.808} & 0.772 & 0.711 &
0.550 & 0.624\\

&
CardioJEPA &
\textbf{0.323} & \textbf{0.702} & \textbf{0.604} & \textbf{0.606} &
\textbf{0.908} & \textbf{0.791} & \textbf{0.699} & \textbf{0.493} &
\textbf{0.741} & 0.807 & \textbf{0.790} & \textbf{0.742} &
\textbf{0.637} & \textbf{0.699}\\

\midrule

\textbf{CMR} &
\textit{CMR Target} &
\textit{0.661} & \textit{0.960} & \textit{0.954} & \textit{0.939} &
\textit{0.528} & \textit{0.235} & \textit{0.167} & \textit{0.028} &
\textit{0.773} & \textit{0.796} & \textit{0.765} & \textit{0.786} &
\textit{0.951} & \textit{0.182}\\
\bottomrule
\end{tabular}
\vspace{-0.5em}
\begin{flushleft}
$\dagger$ ECCL released no checkpoint and uses full fine-tuning; scores are quoted from the original publication. \\
$\ddagger$ MMCL's original publication reports fine-tuning with attention pooling; the released checkpoint is evaluated here under our linear-probing protocol.
\end{flushleft}
\end{table*}

The results in Table~\ref{tab:target} show that static anatomy and cardiac motion provide complementary information. The ED embedding strongly represents ventricular volumes and mass but captures cardiac function poorly, whereas the motion representation ($\bm \Delta)$ substantially improves LVEF and LVGLS prediction. Concatenating the raw CineMA embeddings provides the strongest overall balance across anatomical and functional phenotypes, but its high dimensionality ($d=1536$) makes it impractical as an alignment target. The proposed \textit{isolated-PCA} target separately compresses the anatomy and motion streams into a compact $d=512$ representation, matching the supervised target on anatomical phenotypes while providing stronger functional information. Despite being unsupervised, it matches the supervised alternative on anatomy and exceeds it on function (LVEF $R^2=0.663$ versus $0.628$).

\subsection{The Role of Alignment and Preservation Objectives in Retaining ECG-Unique Information} \label{sec:paradigm}
To isolate the contribution of the alignment objective itself, we compare CardioJEPA against a matched contrastive baseline that differs only in the alignment loss. Specifically, we replace the predictive JEPA objective with an asymmetric InfoNCE objective~\cite{ding2024cross,turgut2025ecgcmr}, while keeping the frozen CMR target, ECG encoder, and optimization protocol identical. For the InfoNCE loss, the embeddings are $\ell_2$-normalized, and the temperature is set to $0.1$.

First, we disable the preservation term ($\alpha=0$) to isolate the effect of the alignment objective from the explicit preservation objective. The results show that both paradigms transfer an equivalent amount of shared structural cardiac information ($\Stil=0.636\pm0.002$ for both). In contrast, CardioJEPA retains approximately 2.8\% more ECG-unique information than contrastive learning ($\Util=0.660\pm0.008$ versus $0.642\pm0.003$; Table~\ref{tab:pure}). These findings indicate that predictive alignment preserves ECG-unique information even without an explicit preservation objective, whereas contrastive alignment suppresses it (Section~\ref{sec:contrastive-limitations}). 

We next quantify the cost of alignment, defined as the reduction in ECG-unique information ($\Util$) relative to the original unaligned ECG representation. The ECG-FM baseline provides this upper bound, attaining $\Util=0.709$ (Table~\ref{tab:cost}). Without explicit preservation, alignment reduces $\Util$ to $0.660$ ($\Delta=-0.049$) for CardioJEPA and $0.642$ ($\Delta=-0.067$) for contrastive learning, indicating that predictive alignment sacrifices less ECG-unique information than its contrastive counterpart. Introducing the preservation objective recovers nearly all of this loss: CardioJEPA with distillation reaches $0.699$ ($\Delta=-0.010$), while contrastive learning with distillation reaches $0.689$ ($\Delta=-0.020$). These results show that explicit preservation largely eliminates the cost of alignment, with predictive alignment remaining consistently closer to the unaligned ECG representation.

\subsection{Latent Space Geometry}\label{sec:latent}
To evaluate how the alignment objective shapes the geometry of the shared latent space, we compare the aligned ECG and CMR embeddings produced by the predictive and contrastive alignment objectives (Fig.~\ref{fig:gap}). Following Liang et al.~\cite{liang2022mindgapunderstandingmodality}, we measure the modality gap as the $\ell_2$ distance between the centroids of $\ell_2$-normalized aligned ECG and CMR embeddings. We report this analysis on the supervised target space.

The unaligned ECG-FM baseline exhibits the largest modality gap ($1.377$). After alignment, CardioJEPA reduces this gap to $0.190$, producing a broad, overlapping latent manifold, whereas contrastive learning retains a substantially larger separation ($0.454$) by concentrating the two modalities into distinct cones. This is consistent with our theoretical analysis: the predictive objective attracts ECG representations toward their CMR targets without the repulsive forces inherent to contrastive learning, so the ECG manifold can mirror the geometry of the frozen target rather than being compressed into a cone. We report this as a descriptive property of the aligned latent space; whether a smaller modality gap has downstream benefit is not tested here.

\subsection{Comparison with Prior Work}\label{sec:positioning}
To position our work against existing ECG--CMR alignment methods, we compare our approach with the MMCL~\cite{turgut2025ecgcmr}, PTACL~\cite{selivanov2026PTACL}, and ECCL~\cite{ding2024cross} frameworks. The ablations in Sections~\ref{sec:paradigm} and \ref{sec:latent} identify CardioJEPA with distillation as the best-performing configuration, which we use for all subsequent experiments. We compare against the released MMCL and PTACL checkpoints under our linear-probing protocol on the held-out split, while reporting ECCL from the original publication because no public checkpoint is available (full fine-tuning, marked~$\dagger$).

CardioJEPA achieves the strongest overall performance among ECG--CMR alignment methods across structural and electrical phenotypes (Table~\ref{tab:positioning}). Notably, the unaligned ECG-FM initialization already exceeds all previously published alignment methods on the four structural phenotypes. CardioJEPA further increases the structural composite to $\tilde S=0.637$, while retaining an ECG-unique composite close to the pre-alignment baseline ($\tilde U_{\mathrm{ecg}}=0.699$ versus $0.709$). Among aligned methods, CardioJEPA preserves the largest amount of the original ECG representation, remaining closest to ECG-FM across all four electrical phenotypes. It also achieves the highest AUROC among aligned methods for atrial fibrillation, cardiomyopathy, and myocardial infarction, while the unaligned ECG-FM remains strongest for bundle branch block. These results demonstrate that CMR-derived structural information can be incorporated into the ECG representation without substantially sacrificing its original electrophysiological content.

\subsection{Out-of-Domain External Validation}\label{sec:ood}
To assess whether the learned representations generalize beyond the UK Biobank training distribution, we evaluate CardioJEPA on two external tasks: structural heart disease detection in EchoNext~\cite{echonext-dataset} and Chagas disease detection across independent Latin American cohorts. \\

\begin{table}[t]
\centering
\caption{Structural heart disease detection (EchoNext~\cite{echonext-dataset}).}
\label{tab:echonext_shd}
\begin{tabular}{lcc}
\toprule
\textbf{Method} & \textbf{AUROC} & \textbf{AUPRC} \\
\midrule
EchoNext$^{\dagger}$ \cite{echonext}& 0.831 & 0.750 \\
EchoNext-mini \cite{hughes2026echonext}& 0.820 [0.809,0.831] & 0.789 [0.772,0.804] \\
CardioJEPA (ours) & \textbf{0.850} [0.839,0.861] & \textbf{0.823} [0.809,0.839] \\
\bottomrule
\end{tabular}
\vspace{-0.5em}
\begin{flushleft}
${\dagger}$ Full EchoNext model, trained on 1.2M private multicentre pairs, evaluated on this public test split (results quoted from \cite{echonext}).
\end{flushleft}
\vspace{-1em}
\end{table}

\subsubsection{Structural heart disease detection} To evaluate whether the structural cardiac information learned through ECG--CMR alignment transfers across datasets and patient populations, we compare CardioJEPA against the EchoNext models~\cite{echonext,hughes2026echonext} on the EchoNext dataset~\cite{echonext-dataset}. EchoNext and EchoNext-mini are state-of-the-art models for detection of structural heart disease (SHD) using ECG waveform, demographics, and ECG-derived tabular features. Following the EchoNext benchmark, we evaluate performance on the primary SHD composite endpoint, which is positive if any of 11 individual structural abnormalities are present. Both EchoNext models are trained end-to-end for this task, whereas our encoder is evaluated in a frozen-representation setting, where only a linear probe is trained on top of the aligned ECG embeddings without any SHD-specific optimization. Despite this asymmetry, CardioJEPA exceeds both end-to-end baselines (Table~\ref{tab:echonext_shd}). Relative to EchoNext-mini, AUROC improves from $0.820$ to $0.850$ and AUPRC from $0.789$ to $0.823$, with non-overlapping confidence intervals in both cases. The margin also holds against the full EchoNext model, trained on 1.2 million private multicentre pairs, which reaches $0.831$ AUROC and $0.750$ AUPRC on this split: CardioJEPA improves on it by $+0.019$ AUROC and $+0.073$ AUPRC. CardioJEPA further exceeds EchoNext-mini on 8 of the 11 individual abnormalities, with the largest margin on reduced ejection fraction (LVEF $\leq 45\%$, $0.912$ versus $0.852$ AUROC). The learned ECG representation therefore captures a broad spectrum of structural cardiac information rather than a single dominant phenotype. \\


\begin{table}
\centering
\caption{Frozen-encoder transfer to Chagas disease classification.}

\label{tab:chagas}
\begin{tabular}{lcc}
\toprule
\textbf{Method} & \textbf{AUROC} &  \textbf{Top5\%-TPR} \\
\midrule
\multicolumn{3}{c}{\textit{CODE-15 + SaMi-Trop + PTB-XL (2025 PhysioNet Challenge split)}} \\
Van Santvliet \textit{et al}.~\cite{team-biomed} (frozen)$^{\dagger}$ & -- & $0.381 \pm 0.003$ \\
ECG-FM (init) & $0.838 \pm 0.006$ & $0.386 \pm 0.015$ \\
CardioJEPA (ours) & $\mathbf{0.859} \pm 0.005$ & $\mathbf{0.432} \pm 0.008$ \\
\midrule
\multicolumn{3}{c}{\textit{CODE-15 + SaMi-Trop (Jidling et al. split)}} \\
Jidling \textit{et al}.~\cite{jidling2023screening} (end-to-end)$^{\ddagger}$ & $0.80$ & -- \\
ECG-FM (init) & $0.831 \pm 0.008$ & $0.378 \pm 0.016$ \\
CardioJEPA (ours) & $\mathbf{0.852} \pm 0.005$ & $\mathbf{0.425} \pm 0.012$ \\
\bottomrule
\end{tabular}
\vspace{-0.5em}
\begin{flushleft}
${\dagger}$ Frozen-backbone ablation of the challenge winners, matched to our frozen setting. Their fine-tuned model reached $0.490$ top-5\% TPR ($0.867$ AUROC). \\
${\ddagger}$ In-distribution validation AUROC; their CNN was trained end-to-end on CODE and SaMi-Trop, whereas our encoder is applied out-of-distribution. 
\end{flushleft}
\vspace{-1em}
\end{table}

\subsubsection{Chagas disease detection} Chagas disease is a parasitic cardiomyopathy whose early, treatable stage is frequently missed. Detecting it from the 12-lead ECG alone was the task of the 2025 George B. Moody PhysioNet Challenge~\cite{chagaschallenge}, and was first approached with deep learning by Jidling \textit{et al.}~\cite{jidling2023screening}, who trained a ResNet ensemble on the CODE and SaMi-Trop cohorts. Chagas cardiomyopathy combines structural myocardial remodeling with conduction and rhythm abnormalities, making it an ideal setting to evaluate both components of the learned representation. Successful transfer requires both the structural cardiac information acquired through ECG--CMR alignment and the electrophysiological information preserved during alignment.

We evaluate the frozen representation using five-fold cross validation under two published protocols (Table~\ref{tab:chagas}), applying the encoder fully out of distribution: it has never seen these Latin American cohorts or any Chagas label during pre-training. Under the Challenge protocol, the unaligned ECG-FM initialization already matches the frozen configuration of the challenge-winning system~\cite{team-biomed} (top-5\% TPR $0.386$ versus $0.381$), confirming that large-scale ECG pretraining alone captures much of the relevant signal. CardioJEPA improves on this foundation rather than simply inheriting it, increasing top-5\% TPR to $0.432$ and AUROC to $0.859$; the same pattern holds under the Jidling \textit{et al}.\ protocol, where our models reach an AUROC of $0.852$, above the $0.80$ reported for an end-to-end CNN trained \emph{in distribution} on these cohorts. 
These results show that an imaging-based, disease-agnostic ECG representation can detect an unseen disease spanning both structural and electrophysiological abnormalities, providing direct evidence that cross-modal alignment does not require sacrificing modality-unique information.

\section{Discussion}\label{sec:discussion}
This study introduced CardioJEPA, a framework for ECG--CMR alignment that combines an informative CMR target representation with a predictive objective and explicit preservation of ECG-unique information. The resulting cardiac representation consistently outperformed previous ECG--CMR alignment methods across structural, electrical, and diagnostic tasks. Notably, the unaligned ECG foundation model already exceeded all of them on structural phenotype regression (Table~\ref{tab:positioning}), highlighting the importance of large-scale unimodal pre-training. This observation is consistent with the recent shift toward large-scale self-supervised foundation models~\cite{vaid2023foundational,coppola2024hubert,li2025ecgfounder}. Building on this initialization, CardioJEPA further improved structural performance while preserving the electrophysiological information that gives the ECG its unique clinical value, indicating that multimodal alignment contributes cardiac information beyond what ECG pre-training alone provides.

The central finding of this work is that incorporating structural information from CMR does not require sacrificing ECG-specific information. As discussed in Section~\ref{sec:contrastive-limitations}, contrastive objectives are theoretically expected to compress modality-specific information. The empirical margin is nonetheless modest, because the bound in Eq.~(\ref{eq:uecg_ceiling}) describes the asymptotic limit of InfoNCE: alignment against a frozen target converges before the ECG encoder reaches minimal sufficiency, so the suppressive effect appears as a small difference rather than the collapse the theory describes in the limit. Our experiments (Table~\ref{tab:pure}) show that, when aligning to a frozen CMR target, both predictive and contrastive objectives transfer a similar amount of structural information. Their primary difference lies in how well they preserve ECG-specific information: predictive alignment consistently retains more, and adding the distillation objective lifts both paradigms close to the pre-alignment ceiling. The resulting representation approaches the original ECG foundation model on electrophysiological phenotypes while simultaneously improving structural prediction, demonstrating that preserving ECG-specific information is not incompatible with learning imaging-derived structural representations. These findings support the theoretical motivation for predictive latent alignment ~\cite{assran2023jepa,perera2025crossjepa} in complementary modalities, while also emphasizing that explicit preservation is an essential component of successful ECG--CMR alignment.

Our experiments further demonstrate that the CMR target representation is equally important as it determines the structural information available for transfer. As motivated in Section~\ref{sec:problem}, alignment cannot recover information absent from the target representation. The ablations in Table~\ref{tab:target} show this directly. A representation derived from a single end-diastolic frame accurately captures ventricular anatomy but poorly represents cardiac function, whereas separating static anatomy from cardiac motion substantially improves functional phenotypes. The proposed isolated-PCA target preserves this complementary information in a compact latent representation and, despite requiring no supervision, matches the supervised alternative on anatomical phenotypes while exceeding it on cardiac function. Overall, this representation provides a compact but efficient solution to maximize the information that is available for transfer. 

Unlike previous ECG--CMR alignment methods, which were evaluated only in-distribution~\cite{ding2024cross,turgut2025ecgcmr,selivanov2026PTACL}, we showed that CardioJEPA generalized across external datasets, downstream tasks, and disease populations. The magnitude of improvement, however, depended on the nature of the target disease. Purely electrophysiological conditions, such as bundle branch block, derived little benefit from imaging supervision because the relevant information is already contained within the ECG (Table~\ref{tab:positioning}). In contrast, diseases characterized by structural remodeling consistently benefited from multimodal alignment. Most notably, despite relying only on a frozen representation and a linear probe, CardioJEPA outperformed the end-to-end EchoNext-mini model~\cite{hughes2026echonext} for structural heart disease detection (Table~\ref{tab:echonext_shd}). It likewise improved ECG-based detection of Chagas cardiomyopathy relative to the frozen configuration of the challenge-winning model~\cite{team-biomed} (Table~\ref{tab:chagas}), without disease-specific fine-tuning. The finding that the same frozen representation supports both an echo-derived structural endpoint and a parasitic cardiomyopathy, in cohorts and populations it was never trained on, suggests it encodes a broad cardiac state rather than any single phenotype.

Several limitations should be acknowledged. 
First, our evaluation relies on frozen representations and linear probing. Electrophysiological information recoverable only through nonlinear adaptation is therefore not captured, and the representational capacity of CardioJEPA may be underestimated. 
Second, the framework is limited to cine CMR and can only transfer information present in that modality; extending the target to sequences that capture tissue characteristics, such as late gadolinium enhancement or parametric mapping, may further enrich the resulting ECG representation. 
Finally, although CardioJEPA generalized across multiple external datasets and clinical tasks, prospective multicenter validation is required to determine whether these improvements directly translate into earlier diagnosis in clinical practice~\cite{kemper2025esr,ramwala2024establishing}.

\section{Conclusion}\label{sec:conclusion}
We presented CardioJEPA, a predictive ECG--CMR alignment framework built on frozen cardiac foundation models. 
Comparing target construction, alignment, and preservation under one protocol shows that the CMR target and the preservation term govern downstream performance, while the two alignment objectives, under an identical target and encoder, differ mainly in how much ECG-unique information they retain, and in the geometry of the resulting latent space. CardioJEPA outperforms published ECG--CMR alignment methods, and the predictive objective yields a more coherent shared latent space than a matched contrastive control. The resulting frozen representations detect structural heart disease and Chagas cardiomyopathy in external cohorts without task-specific training, suggesting that ECG models carrying imaging-derived structure could extend cardiac assessment where CMR is unavailable.

\section*{REFERENCES}
\bibliographystyle{IEEEtran}
\bibliography{ref}

@IEEEtranBSTCTL{IEEEexample:BSTcontrol,
  CTLuse_forced_etal       = "yes",
  CTLmax_names_forced_etal = "6",
  CTLnames_show_etal       = "1"
}

@article{frisch1933regression,
 ISSN = {00129682, 14680262},
 author = {Ragnar Frisch and Frederick V. Waugh},
 journal = {Econometrica},
 number = {4},
 pages = {387--401},
 publisher = {[Wiley, Econometric Society]},
 title = {{Partial Time Regressions as Compared with Individual Trends}},
 volume = {1},
 year = {1933}
}

@article{kemper2025esr,
  title={{ESR Essentials: how to get to valuable radiology AI: the role of early health technology assessment—practice recommendations by the European Society of Medical Imaging Informatics}},
  author={Kemper, Erik HM and Erenstein, Hendrik and Boverhof, Bart-Jan and Redekop, Ken and Andreychenko, Anna E and Dietzel, Matthias and Groot Lipman, Kevin BW and Huisman, Merel and Klontzas, Michail E and Vos, Frans and others},
  journal={European radiology},
  volume={35},
  number={6},
  pages={3432--3441},
  year={2025},
  publisher={Springer}
}

@article{ramwala2024establishing,
  title={{Establishing a validation infrastructure for imaging-based artificial intelligence algorithms before clinical implementation}},
  author={Ramwala, Ojas A and Lowry, Kathryn P and Cross, Nathan M and Hsu, William and Austin, Christopher C and Mooney, Sean D and Lee, Christoph I},
  journal={Journal of the American College of Radiology},
  volume={21},
  number={10},
  pages={1569--1574},
  year={2024},
  publisher={Elsevier}
}

@article{echonext,
  title={{Detecting structural heart disease from electrocardiograms using AI}},
  author={Poterucha, Timothy J and Jing, Linyuan and Ricart, Ramon Pimentel and Adjei-Mosi, Michael and Finer, Joshua and Hartzel, Dustin and Kelsey, Christopher and Long, Aaron and Rocha, Daniel and Ruhl, Jeffrey A and others},
  journal={Nature},
  volume={644},
  number={8075},
  pages={221--230},
  year={2025},
  publisher={Nature Publishing Group UK London}
}

@article{hughes2026echonext,
  title={{EchoNext-mini: A dataset and baseline AI model for detecting structural heart disease from electrocardiograms}},
  author={Hughes, John Weston and Jing, Linyuan and Finer, Joshua and Hartzel, Dustin and Kelsey, Christopher and Long, Aaron and Rocha, Daniel and Ruhl, Jeffrey and Poterucha, Timothy and Elias, Pierre},
  journal={{NEJM AI}},
  volume={3},
  number={5},
  pages={AIdbp2500516},
  year={2026},
  publisher={Massachusetts Medical Society}
}

@article{pennell2010cardiovascular,
  title={{Cardiovascular magnetic resonance}},
  author={Pennell, Dudley J},
  journal={Circulation},
  volume={121},
  number={5},
  pages={692--705},
  year={2010},
  publisher={Lippincott Williams \& Wilkins}
}

@article{echonext-dataset,
  author = {Elias, Pierre and Finer, Joshua},
  title = {{EchoNext: A Dataset for Detecting Echocardiogram-Confirmed Structural Heart Disease from ECGs}},
  journal = {{PhysioNet}},
  year = {2026},
  month = apr,
  note = {Version 1.1.1},
  doi = {10.13026/7cfw-d091},
  url = {https://doi.org/10.13026/7cfw-d091}
}

@article{chaitanya2020contrastive,
  title={{Contrastive learning of global and local features for medical image segmentation with limited annotations}},
  author={Chaitanya, Krishna and Erdil, Ertunc and Karani, Neerav and Konukoglu, Ender},
  journal={Advances in neural information processing systems},
  volume={33},
  pages={12546--12558},
  year={2020}
}

@article{li2025ecgfounder,
  title={{An electrocardiogram foundation model built on over 10 million recordings}},
  author={Li, Jun and Aguirre, Aaron D and Junior, Valdery Moura and Jin, Jiarui and Liu, Che and Zhong, Lanhai and Sun, Chenxi and Clifford, Gari and Brandon Westover, M and Hong, Shenda},
  journal={{NEJM AI}},
  volume={2},
  number={7},
  pages={AIoa2401033},
  year={2025},
  publisher={Massachusetts Medical Society}
}

@article{coppola2024hubert,
  title={{HuBERT-ECG as a self-supervised foundation model for broad and scalable cardiac applications}},
  author={Coppola, Edoardo and Savardi, Mattia and Massussi, Mauro and Adamo, Marianna and Metra, Marco and Signoroni, Alberto},
  journal={{medRxiv}},
  year={2024},
  publisher={Cold Spring Harbor Laboratory Press}
}

@article{vaid2023foundational,
  title={{A foundational vision transformer improves diagnostic performance for electrocardiograms}},
  author={Vaid, Akhil and Jiang, Joy and Sawant, Ashwin and Lerakis, Stamatios and Argulian, Edgar and Ahuja, Yuri and Lampert, Joshua and Charney, Alexander and Greenspan, Hayit and Narula, Jagat and others},
  journal={NPJ digital medicine},
  volume={6},
  number={1},
  pages={108},
  year={2023},
  publisher={Nature Publishing Group UK London}
}

@article{kligfield2007recommendations,
  title={{Recommendations for the standardization and interpretation of the electrocardiogram}},
  author={Kligfield, Paul and Gettes, Leonard S and Bailey, James J and Childers, Rory and Deal, Barbara J and Hancock, E William and Van Herpen, Gerard and Kors, Jan A and Macfarlane, Peter and Mirvis, David M and others},
  journal={Circulation},
  volume={115},
  number={10},
  pages={1306--1324},
  year={2007},
  publisher={Lippincott Williams \& Wilkins}
}

@article{jidling2023screening,
  title={{Screening for Chagas disease from the electrocardiogram using a deep neural network}},
  author={Jidling, Carl and Gedon, Daniel and Sch{\"o}n, Thomas B and Oliveira, Claudia Di Lorenzo and Cardoso, Clareci Silva and Ferreira, Ariela Mota and Giatti, Luana and Barreto, Sandhi Maria and Sabino, Ester C and Ribeiro, Antonio LP and others},
  journal={PLoS neglected tropical diseases},
  volume={17},
  number={7},
  pages={e0011118},
  year={2023},
  publisher={Public Library of Science San Francisco, CA USA}
}

@article{team-biomed,
  title={{Detecting Chagas disease using a vision transformer-based ECG foundation model}},
  author={Van Santvliet, Lore and Nguyen, Phu Xuan and Vandenberk, Bert and De Vos, Maarten},
  journal={Proceedings of CinC 2025},
  volume={52},
  year={2025}
}

@inproceedings{chagaschallenge,
author = {Reyna, Matthew A. and Koscova, Zuzana and Pavlus, Jan and Weigle, James and Saghafi, Soheil and Gomes, Paulo and Elola, Andoni and Hassannia, Mohammad Sina and Campbell, Kiersten and Bahrami Rad, Ali and Ribeiro, Antônio H and Ribeiro, Antonio LP and Sameni, Reza and Clifford, Gari D},
title = {{Detection of Chagas Disease from the ECG: The George B. Moody PhysioNet Challenge 2025}},
booktitle={{Computing in Cardiology 2025}},
volume={52},
pages={1--4},
year={2025}}

@inproceedings{liang2022mindgapunderstandingmodality,
 author = {Liang, Victor Weixin and Zhang, Yuhui and Kwon, Yongchan and Yeung, Serena and Zou, James},
 booktitle = {Advances in Neural Information Processing Systems},
 doi = {10.52202/068431-1280},
 editor = {S. Koyejo and S. Mohamed and A. Agarwal and D. Belgrave and K. Cho and A. Oh},
 pages = {17612--17625},
 publisher = {Curran Associates, Inc.},
 title = {{Mind the Gap: Understanding the Modality Gap in Multi-modal Contrastive Representation Learning}},
 volume = {35},
 year = {2022}
}

@inproceedings{chuang2020debiased,
 author = {Chuang, Ching-Yao and Robinson, Joshua and Lin, Yen-Chen and Torralba, Antonio and Jegelka, Stefanie},
 booktitle = {{Advances in Neural Information Processing Systems}},
 editor = {H. Larochelle and M. Ranzato and R. Hadsell and M.F. Balcan and H. Lin},
 pages = {8765--8775},
 publisher = {Curran Associates, Inc.},
 title = {{Debiased Contrastive Learning}},
 volume = {33},
 year = {2020}
}

@article{oord2019representationlearningcontrastivepredictive,
      title={{Representation Learning with Contrastive Predictive Coding}}, 
      author={Aaron van den Oord and Yazhe Li and Oriol Vinyals},
      journal = {arXiv preprint arXiv:1807.03748},
      year={2018},
      eprint={1807.03748},
      archivePrefix={arXiv},
      primaryClass={cs.LG},
}

@article{ding2026generating,
author = {Zhengyao Ding  and Ziyu Li  and Yujian Hu  and Youyao Xu  and Chengchen Zhao  and Yiheng Mao  and Haitao Li  and Zhikang Li  and Qian Li  and Jing Wang  and Yue Chen  and Mengjia Chen  and Longbo Wang  and Xuesen Chu  and Weichao Pan  and Ziyi Liu  and Fei Wu  and Hongkun Zhang  and Ting Chen  and Zhengxing Huang },
title = {{Generating Cardiac Magnetic Resonance Images from Electrocardiograms — A Multicenter Study}},
journal = {{NEJM AI}},
volume = {3},
number = {4},
pages = {AIoa2500549},
year = {2026},
doi = {10.1056/AIoa2500549},

eprint = {https://ai.nejm.org/doi/pdf/10.1056/AIoa2500549}
}

@article{ribeiro_automatic_2020,
  title = {{Automatic Diagnosis of the 12-Lead {{ECG}} Using a Deep Neural Network}},
  author = {Ribeiro, Ant{\^o}nio H. and Ribeiro, Manoel Horta and Paix{\~a}o, Gabriela M. M. and Oliveira, Derick M. and Gomes, Paulo R. and Canazart, J{\'e}ssica A. and Ferreira, Milton P. S. and Andersson, Carl R. and Macfarlane, Peter W. and Meira Jr., Wagner and Sch{\"o}n, Thomas B. and Ribeiro, Antonio Luiz P.},
  year = {2020},
  volume = {11},
  pages = {1760},
  doi = {https://doi.org/10.1038/s41467-020-15432-4},
  journal = {Nature Communications},
  number = {1}
}

@article{oh2018automated,
title = {{Automated diagnosis of arrhythmia using combination of CNN and LSTM techniques with variable length heart beats}},
journal = {Computers in Biology and Medicine},
volume = {102},
pages = {278-287},
year = {2018},
issn = {0010-4825},
doi = {https://doi.org/10.1016/j.compbiomed.2018.06.002},
author = {Shu Lih Oh and Eddie Y.K. Ng and Ru San Tan and U. Rajendra Acharya}
}

@inproceedings{tian2020infomin,
 author = {Tian, Yonglong and Sun, Chen and Poole, Ben and Krishnan, Dilip and Schmid, Cordelia and Isola, Phillip},
 booktitle = {{Advances in Neural Information Processing Systems}},
 editor = {H. Larochelle and M. Ranzato and R. Hadsell and M.F. Balcan and H. Lin},
 pages = {6827--6839},
 publisher = {Curran Associates, Inc.},
 title = {{What Makes for Good Views for Contrastive Learning?}},
 volume = {33},
 year = {2020}
}

@inproceedings{wang2020hypersphere,
  title={{Understanding Contrastive Representation Learning through Alignment and Uniformity on the Hypersphere}},
  author={Wang, Tongzhou and Isola, Phillip},
  booktitle={{International Conference on Machine Learning}},
  organization={PMLR},
  pages={9929--9939},
  year={2020}
}

@inproceedings{lee2021predicting,
 author = {Lee, Jason D and Lei, Qi and Saunshi, Nikunj and Zhuo, Jiacheng},
 booktitle = {{Advances in Neural Information Processing Systems}},
 editor = {M. Ranzato and A. Beygelzimer and Y. Dauphin and P.S. Liang and J. Wortman Vaughan},
 pages = {309--323},
 publisher = {Curran Associates, Inc.},
 title = {{Predicting What You Already Know Helps: Provable Self-Supervised Learning}},
 volume = {34},
 year = {2021}
}

@InProceedings{tosh2021contrastive,
  title = 	 {Contrastive learning, multi-view redundancy, and linear models},
  author =       {Tosh, Christopher and Krishnamurthy, Akshay and Hsu, Daniel},
  booktitle = 	 {Proceedings of the 32nd International Conference on Algorithmic Learning Theory},
  pages = 	 {1179--1206},
  year = 	 {2021},
  editor = 	 {Feldman, Vitaly and Ligett, Katrina and Sabato, Sivan},
  volume = 	 {132},
  series = 	 {Proceedings of Machine Learning Research},
  month = 	 {16--19 Mar},
  publisher =    {PMLR}
}

@INPROCEEDINGS {wang2022rethinking,
author = { Wang, Haoqing and Guo, Xun and Deng, ZhiHong and Lu, Yan },
booktitle = {{2022 IEEE/CVF Conference on Computer Vision and Pattern Recognition (CVPR)}},
title = {{Rethinking Minimal Sufficient Representation in Contrastive Learning}},
year = {2022},
volume = {},
ISSN = {},
pages = {16020-16029},
doi = {10.1109/CVPR52688.2022.01557},
publisher = {IEEE Computer Society},
address = {Los Alamitos, CA, USA},
month =Jun}

@article{lecun2022path,
  title={{A Path Towards Autonomous Machine Intelligence}},
  author={LeCun, Yann},
  journal={OpenReview},
  year={2022},
}

@article{perera2025crossjepa,
      title={{CrossJEPA: Cross-Modal Joint-Embedding Predictive Architecture for Efficient 3D Representation Learning from 2D Images}}, 
      author={Avishka Perera and Kumal Hewagamage and Saeedha Nazar and Kavishka Abeywardana and Hasitha Gallella and Ranga Rodrigo and Mohamed Afham},
      year={2025},
      journal = {arXiv preprint arXiv:2511.18424},
      eprint={2511.18424},
      archivePrefix={arXiv},
      primaryClass={cs.CV},
}

@InProceedings{assran2023jepa,
    author    = {Assran, Mahmoud and Duval, Quentin and Misra, Ishan and Bojanowski, Piotr and Vincent, Pascal and Rabbat, Michael and LeCun, Yann and Ballas, Nicolas},
    title     ={{Self-Supervised Learning From Images With a Joint-Embedding Predictive Architecture}},
    booktitle = {{Proceedings of the IEEE/CVF Conference on Computer Vision and Pattern Recognition (CVPR)}},
    month     = {June},
    year      = {2023},
    pages     = {15619-15629}
}

@inproceedings{liang2023factorized,
 author = {Liang, Paul Pu and Deng, Zihao and Ma, Martin Q. and Zou, James and Morency, Louis-Philippe and Salakhutdinov, Ruslan},
 booktitle = {{Advances in Neural Information Processing Systems}},
 doi = {10.52202/075280-1430},
 editor = {A. Oh and T. Naumann and A. Globerson and K. Saenko and M. Hardt and S. Levine},
 pages = {32971--32998},
 publisher = {Curran Associates, Inc.},
 title = {{Factorized Contrastive Learning: Going Beyond Multi-view Redundancy}},
 volume = {36},
 year = {2023}
}

@inproceedings{ronneberger2015unet,
  title={{U-net: Convolutional networks for biomedical image segmentation}},
  author={Ronneberger, Olaf and Fischer, Philipp and Brox, Thomas},
  booktitle={{International Conference on Medical image computing and computer-assisted intervention}},
  pages={234--241},
  year={2015},
  organization={Springer}
}

@INPROCEEDINGS{gedon2021towardssslecg,
  author={Gedon, Daniel and Ribeiro, Antônio H. and Wahlström, Niklas and Schön, Thomas B.},
  booktitle={{2021 Computing in Cardiology (CinC)}}, 
  title={{First Steps Towards Self-Supervised Pretraining of the 12-Lead ECG}}, 
  year={2021},
  volume={48},
  number={},
  pages={1-4},
  doi={10.23919/CinC53138.2021.9662748}}

@article{mehari2021sslecg,
author = {Mehari, Temesgen and Strodthoff, Nils},
year = {2021},
month = {12},
pages = {105114},
title = {{Self-supervised representation learning from 12-lead ECG data}},
volume = {141},
journal = {Computers in Biology and Medicine},
doi = {10.1016/j.compbiomed.2021.105114}
}

@InProceedings{chen2020simclr,
  title = 	 {{A Simple Framework for Contrastive Learning of Visual Representations}},
  author =       {Chen, Ting and Kornblith, Simon and Norouzi, Mohammad and Hinton, Geoffrey},
  booktitle = 	 {{Proceedings of the 37th International Conference on Machine Learning}},
  pages = 	 {1597--1607},
  year = 	 {2020},
  editor = 	 {III, Hal Daumé and Singh, Aarti},
  volume = 	 {119},
  series = 	 {Proceedings of Machine Learning Research},
  month = 	 {13--18 Jul},
  publisher =    {PMLR}
}

@inproceedings{khosla2020supcon,
 author = {Khosla, Prannay and Teterwak, Piotr and Wang, Chen and Sarna, Aaron and Tian, Yonglong and Isola, Phillip and Maschinot, Aaron and Liu, Ce and Krishnan, Dilip},
 booktitle = {{Advances in Neural Information Processing Systems}},
 editor = {H. Larochelle and M. Ranzato and R. Hadsell and M.F. Balcan and H. Lin},
 pages = {18661--18673},
 publisher = {Curran Associates, Inc.},
 title = {{Supervised Contrastive Learning}},
 volume = {33},
 year = {2020}
}

@InProceedings{radford2021clip,
  title = 	 {{Learning Transferable Visual Models From Natural Language Supervision}},
  author =       {Radford, Alec and Kim, Jong Wook and Hallacy, Chris and Ramesh, Aditya and Goh, Gabriel and Agarwal, Sandhini and Sastry, Girish and Askell, Amanda and Mishkin, Pamela and Clark, Jack and Krueger, Gretchen and Sutskever, Ilya},
  booktitle = 	 {{Proceedings of the 38th International Conference on Machine Learning}},
  pages = 	 {8748--8763},
  year = 	 {2021},
  editor = 	 {Meila, Marina and Zhang, Tong},
  volume = 	 {139},
  month = 	 {18--24 Jul},
  publisher =    {PMLR}
}

@inproceedings{ding2024cross,
  title={{Cross-Modality Cardiac Insight Transfer: A Contrastive Learning Approach to Enrich ECG with CMR Features}},
  author={Ding, Zhengyao and Hu, Yujian and Li, Ziyu and Zhang, Hongkun and Wu, Fei and Xiang, Yilang and Li, Tian and Liu, Ziyi and Chu, Xuesen and Huang, Zhengxing},
  booktitle={{International Conference on Medical Image Computing and Computer-Assisted Intervention}},
  pages={109--119},
  year={2024},
  organization={Springer}
}

@inproceedings{selivanov2026PTACL,
title = {{Global and Local Contrastive Learning for Joint Representations from Cardiac MRI and ECG}},
author = "Alexander Selivanov and Philip M{\"u}ller and {\"O}zg{\"u}n Turgut and Nil Stolt-Ans{\'o} and Daniel Rueckert",
year = "2026",
doi = "10.1007/978-3-032-04927-8\_21",
language = "English",
isbn = "9783032049261",
pages = "217--227",
booktitle = {{Medical Image Computing and Computer Assisted Intervention, MICCAI 2025 - 28th International Conference, 2025, Proceedings}},
}

@article{radhakrishnan2023cross,
  title={{Cross-modal autoencoder framework learns holistic representations of cardiovascular state}},
  author={Radhakrishnan, Adityanarayanan and Friedman, Sam F and Khurshid, Shaan and Ng, Kenney and Batra, Puneet and Lubitz, Steven A and Philippakis, Anthony A and Uhler, Caroline},
  journal={Nature Communications},
  volume={14},
  number={1},
  pages={2436},
  year={2023},
  publisher={Nature Publishing Group UK London}
}

@inproceedings{alvarezflorez2026dualphasecrossmodalcontrastivelearning,
  author    = {Laura Alvarez-Florez and Angel Bujalance-Gomez and
               Femke Raijmakers and Samuel Ruiperez-Campillo and
               Maarten Z. H. Kolk and Jesse Wiers and
               Julia E. Vogt and Erik J. Bekkers and
               Ivana I{\v{s}}gum and Fleur V. Y. Tjong},
  title     = {Dual-Phase Cross-Modal Contrastive Learning for CMR-Guided ECG Representations for Cardiovascular Disease Assessment},
  booktitle = {SPIE Medical Imaging 2026: Image Processing},
  year      = {2026}
}

@article{turgut2025ecgcmr,
title = {{Unlocking the diagnostic potential of electrocardiograms through information transfer from cardiac magnetic resonance imaging}},
journal = {Medical Image Analysis},
volume = {101},
pages = {103451},
year = {2025},
issn = {1361-8415},
doi = {https://doi.org/10.1016/j.media.2024.103451},
author = {{\"O}zg{\"u}n Turgut and Philip M{\"u}ller and Paul Hager and Suprosanna Shit and Sophie Starck and Martin J. Menten and Eimo Martens and Daniel Rueckert}
}

@article{fu2025cinema,
  title={{A versatile foundation model for cine cardiac magnetic resonance image analysis tasks}},
  author={Fu, Yunguan and Bai, Wenjia and Yi, Weixi and Manisty, Charlotte and Bhuva, Anish N and Treibel, Thomas A and Moon, James C and Clarkson, Matthew J and Davies, Rhodri Huw and Hu, Yipeng},
  journal={arXiv preprint arXiv:2506.00679},
  year={2025}
}

@InProceedings{He_2022_MAE,
    author    = {He, Kaiming and Chen, Xinlei and Xie, Saining and Li, Yanghao and Doll\'ar, Piotr and Girshick, Ross},
    title     = {{Masked Autoencoders Are Scalable Vision Learners}},
    booktitle = {{Proceedings of the IEEE/CVF Conference on Computer Vision and Pattern Recognition (CVPR)}},
    month     = {June},
    year      = {2022},
    pages     = {16000-16009}
}

@article{mckeen2025ecgfm,
    author = {McKeen, Kaden and Masood, Sameer and Toma, Augustin and Rubin, Barry and Wang, Bo},
    title = {{ECG-FM: an open electrocardiogram foundation model}},
    journal = {JAMIA Open},
    volume = {8},
    number = {5},
    pages = {ooaf122},
    year = {2025},
    month = {10},
    issn = {2574-2531},
    doi = {10.1093/jamiaopen/ooaf122},
    eprint = {https://academic.oup.com/jamiaopen/article-pdf/8/5/ooaf122/64719197/ooaf122.pdf},
}

@article{ukbb,
    doi = {10.1371/journal.pmed.1001779},
    author = {Sudlow, Cathie AND Gallacher, John AND Allen, Naomi AND Beral, Valerie AND Burton, Paul AND Danesh, John AND Downey, Paul AND Elliott, Paul AND Green, Jane AND Landray, Martin AND Liu, Bette AND Matthews, Paul AND Ong, Giok AND Pell, Jill AND Silman, Alan AND Young, Alan AND Sprosen, Tim AND Peakman, Tim AND Collins, Rory},
    journal = {PLOS Medicine},
    publisher = {Public Library of Science},
    title = {{UK Biobank: An Open Access Resource for Identifying the Causes of a Wide Range of Complex Diseases of Middle and Old Age}},
    year = {2015},
    month = {03},
    volume = {12},
    pages = {1-10},
    number = {3},

}

\end{document}